\documentclass[final]{cvpr}

\usepackage{times}
\usepackage{epsfig}
\usepackage{helvet}
\usepackage{courier}
\usepackage{floatrow}
\usepackage{multirow}
\newfloatcommand{capbtabbox}{table}[][\FBwidth]
\usepackage{comment}
\usepackage[utf8]{inputenc}
\usepackage[T1]{fontenc}
\usepackage{amsfonts}
\usepackage{units}
\usepackage{microtype}
\usepackage[dvipsnames]{xcolor}
\usepackage{soul}
\usepackage{url}
\usepackage[small]{caption}
\usepackage{graphicx}
\usepackage{amsmath}
\usepackage{booktabs}
\usepackage{epstopdf}
\usepackage{amsthm}
\usepackage{amssymb}
\usepackage{subfigure}
\usepackage{wrapfig}
\newcommand{\tabincell}[2]{\begin{tabular}{@{}#1@{}}#2\end{tabular}}

\usepackage{bbding}
\usepackage{enumitem}
\usepackage{float}

\usepackage[ruled,vlined]{algorithm2e}
\usepackage{algpseudocode}
\renewcommand{\algorithmicrequire}{\textbf{Input:}} 
\renewcommand{\algorithmicensure}{\textbf{Output:}} 

\usepackage[pagebackref=true,breaklinks=true,colorlinks,bookmarks=false]{hyperref}

\def\cvprPaperID{1442} 
\def\confYear{CVPR 2021}

\begin{document}

\title{Diversifying Sample Generation for Accurate Data-Free Quantization}

\vspace{-0.1in}

\author{
\fontsize{11.0pt}
{\baselineskip}\selectfont Xiangguo Zhang \thanks{Equal contribution.} \textsuperscript{ 1}, 
Haotong Qin\textsuperscript{* 1,5}, Yifu Ding\textsuperscript{1}, Ruihao Gong\textsuperscript{3, 4},\\
Qinghua Yan\textsuperscript{1}, Renshuai Tao\textsuperscript{1}, Yuhang Li\textsuperscript{2}, Fengwei Yu\textsuperscript{3, 4}, Xianglong Liu\textsuperscript{1\thanks{Corresponding author}}\\
\textsuperscript{1}{\fontsize{11.0pt}{\baselineskip}\selectfont Beihang University}\quad
\textsuperscript{2}{\fontsize{11.0pt}{\baselineskip}\selectfont Yale University}\quad
\textsuperscript{3}{\fontsize{11.0pt}{\baselineskip}\selectfont SenseTime Research}\quad
\textsuperscript{4}{\fontsize{11.0pt}{\baselineskip}\selectfont Shanghai AI Laboratory}\\
\textsuperscript{5}{\fontsize{11.0pt}{\baselineskip}\selectfont Shen Yuan Honors College, Beihang University}\\
{\fontsize{8.5pt}{\baselineskip}\selectfont \tt \{xiangguozhang, zjdyf, yanqh, rstao\}@buaa.edu.cn, yuhang.li@yale.edu,}\\
{\fontsize{8.5pt}{\baselineskip}\selectfont \tt \{qinhaotong, xlliu\}@nlsde.buaa.edu.cn, \{gongruihao, yufengwei\}@sensetime.com}
}
\maketitle
\vspace{-0.1in}

\pagestyle{empty}
\thispagestyle{empty}
\begin{abstract}
Quantization has emerged as one of the most prevalent approaches to compress and accelerate neural networks.
Recently, data-free quantization has been widely studied as a practical and promising solution. It synthesizes data for calibrating the quantized model according to the batch normalization (BN) statistics of FP32 ones and significantly relieves the heavy dependency on real training data in traditional quantization methods.
Unfortunately, we find that in practice, the synthetic data identically constrained by BN statistics suffers serious homogenization at both distribution level and sample level and further causes a significant performance drop of the quantized model.
We propose \textbf{D}iverse \textbf{S}ample \textbf{G}eneration (DSG) scheme to mitigate the adverse effects caused by homogenization.
Specifically, we slack the alignment of feature statistics in the BN layer to relax the constraint at the distribution level and design a layerwise enhancement to reinforce specific layers for different data samples.
Our DSG scheme is versatile and even able to be applied to the state-of-the-art post-training quantization method like AdaRound. We evaluate the DSG scheme on the large-scale image classification task and consistently obtain significant improvements over various network architectures and quantization methods, especially when quantized to lower bits (\eg, up to 22\% improvement on W4A4).
Moreover, benefiting from the enhanced diversity, models calibrated with synthetic data perform close to those calibrated with real data and even outperform them on W4A4.
\end{abstract}

\section{Introduction}

\begin{figure}[t]
\centering
\vspace{-0.15in}
\subfigure[]{
\includegraphics[width=0.46\linewidth]{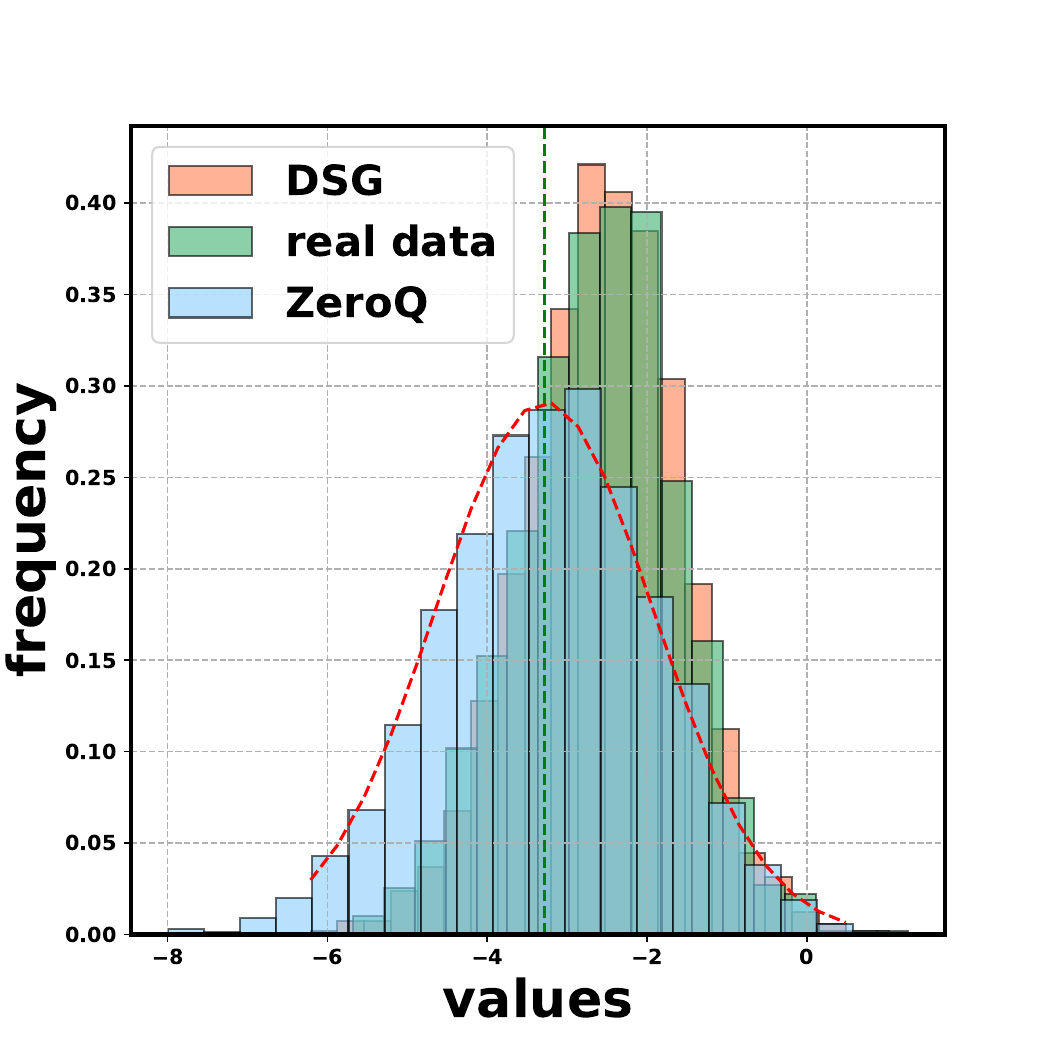}
\label{fig:distribution_a}
}
\hspace{-0.4cm}
\subfigure[]{
\includegraphics[width=0.46\linewidth]{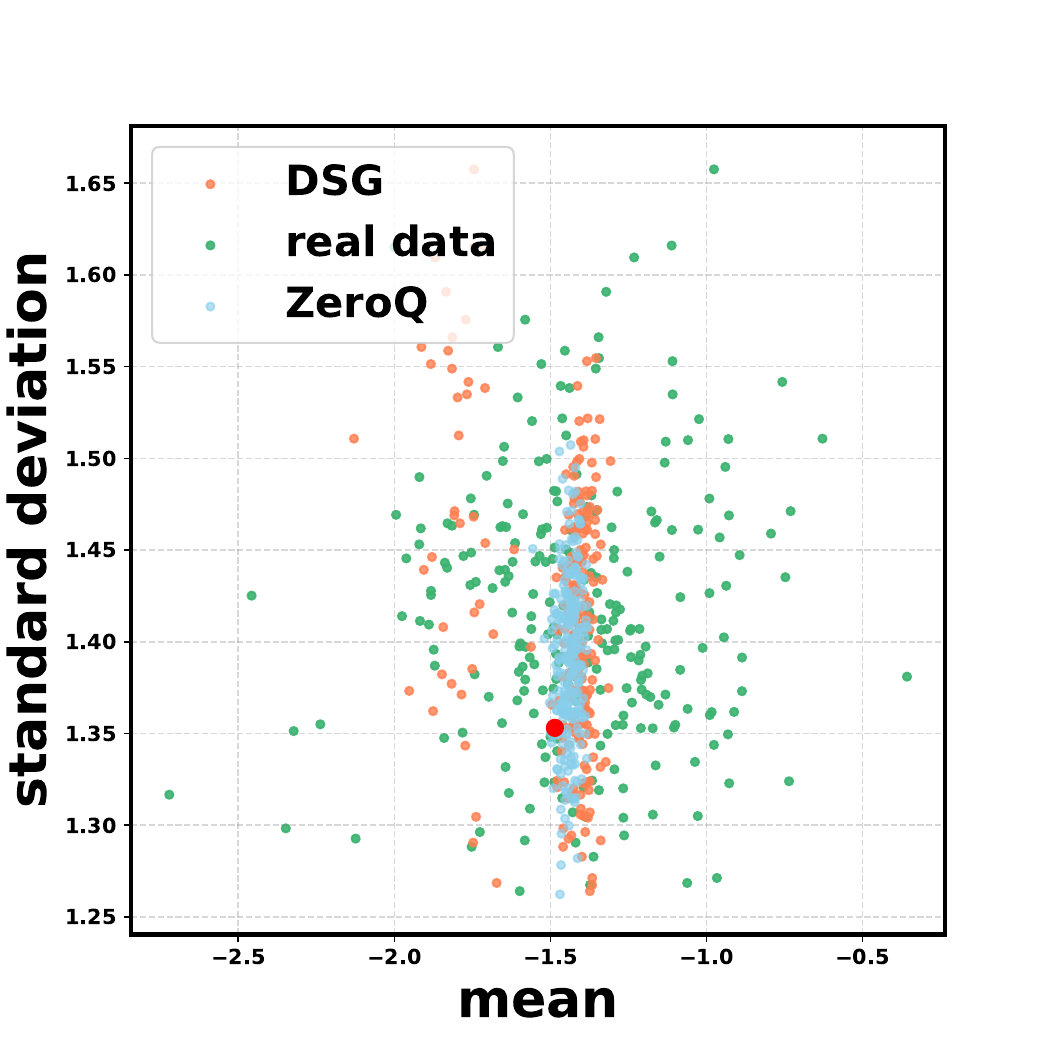}
\label{fig:distribution_b}
}
\vspace{-0.2in}
\caption{Comparison between real data and synthetic data (generated by DSG and ZeroQ~\cite{cai2020zeroq}) with 256 samples of each. (a) shows the activation distribution of one channel in ResNet-18~\cite{he2016deep}. 
ZeroQ data mostly fits the normal distribution of BN statistics (the red line), while real data and DSG data have an offset.
(b) is a chart of the mean and standard deviation of one channel. ZeroQ data gathers near BN statistics (the red dot) but real data and DSG data are more scattered.
}
\label{fig:distribution}
\end{figure}

Recently, Deep Neural Networks (DNNs), especially Convolutional Neural Networks (CNNs) achieve great success in a variety of domains, such as image classification~\cite{krizhevsky2012imagenet,VeryDeepConvolutional,7298594,wang2019dynamic,Wang_2019_ICCV,Yang_2020_CVPR,Wu_2020_CVPR,wang2021dual}, object detection~\cite{DBLP:journals/corr/GirshickDDM13,DBLP:journals/corr/Girshick15,DBLP:journals/corr/abs-1904-02701,NIPS2015_5638,Li_2019_CVPR,WeiOccluded2020}, semantic segmentation~\cite{Everingham:2010:PVO:1747084.1747104,Zhuang_2019_CVPR}, \etc.
Nevertheless, deploying state-of-the-art models on resource-constrained devices is still challenging due to massive parameters and high computational complexity. With more and more hardware support low-precision computations~\cite{Zhu_2020_CVPR,Qin_2020_CVPR,Qin_2020_pr}, quantization has emerged as one of the most promising approaches to obtain efficient neural networks.
Since the whole training stage is required, quantization-aware training methods are considered to be time-consuming and computationally intensive~\cite{gupta2015deep,jacob2018quantization,qin2020bipointnet}. Therefore, post-training quantization methods are proposed, which directly quantize the FP32 models without retraining or fine-tuning~\cite{banner2019posttraining,choukroun2019low,zhao2019improving,nagel2020down, li2021brecq}.
Nevertheless, they still require real training data to calibrate quantized models that is not often ready-to-use for privacy or security concerns, such as medical data and user data.

Fortunately, recent work have proposed data-free quantization to quantize models without any access to real data.
Existing data-free quantization methods~\cite{Nagel_2019_ICCV, cai2020zeroq, haroush2020knowledge, choi2020datafree}, such as ZeroQ~\cite{cai2020zeroq}, generate "optimal synthetic data", which learns an input data distribution to best match the Batch Normalization statistics of the FP32 model.
However, models calibrated with real data perform better than those calibrated with synthetic data, though the synthetic data matches BN statistics better. 
Our study reveals that the data generation process in typical data-free quantization methods has significant homogenization issues at both distribution and sample levels, which prevent models from higher accuracy. 
\textbf{First}, since the synthetic data is constrained to match the BN statistics, the feature distribution in each layer might overfit the BN statistics when data is fed forward in neural networks. 
As shown in Figure~\ref{fig:distribution_a}, the distribution of the synthetic samples generated by ZeroQ almost fits the normal distribution obeying the corresponding BN statistics,
while those of real data have an obvious offset leading to more diverse distribution. We call the phenomenon of the synthetic samples as the distribution level homogenization. 
\textbf{Second}, all samples of synthetic data are optimized by the same objective function in existing generative data-free quantization methods. For instance, ZeroQ and GDFQ~\cite{xu2020generative} apply the same constraint to all data samples, which directly sums the loss objective (KL loss or statistic loss) of each layer. 
Therefore, as shown in Figure~\ref{fig:distribution_b}, the feature distribution statistics of these samples are similar. Specifically, the distribution statistics of synthetic data generated by existing methods are centralized while those of real data are dispersed, as so-called sample level homogenization. 

To mitigate the adverse effects caused by these issues, we propose a novel \textbf{D}iverse \textbf{S}ample \textbf{G}eneration (DSG) scheme, a simple yet effective data generation method for data-free quantization to enhance the diversity of data. 
Our DSG scheme consists of two technical contributions: (1) \textit{Slack Distribution Alignment} (SDA): slack the alignment of the feature statistics in BN layers to relax the constraint of distribution; (2) \textit{Layerwise Sample Enhancement} (LSE): apply the layerwise enhancement to reinforce specific layers for different data samples. 

Our DSG scheme presents a novel perspective of data diversity for data-free quantization.
We evaluate the DSG scheme on the large-scale image classification task, and the results show that DSG performs remarkably well across various network architectures such as ResNet-18~\cite{he2016deep}, ResNet-50, SqueezeNext~\cite{gholami2018squeezenext}, ShuffleNet~\cite{zhang2017shufflenet}, and InceptionV3~\cite{szegedy2016rethinking}, and surpasses previous methods by a wide margin, even outperforms models calibrated with real data. Moreover, we show that the synthetic data generated by DSG scheme can be applied to the most advanced post-training quantization methods, such as AdaRound~\cite{nagel2020down}.

We summarize our main contributions as: 
\begin{enumerate}[nosep, leftmargin=*]
    \item We revisit the data generation process of data-free quantization methods from the diversity perspective. Our study reveals the homogenization problems of synthetic data existing at two levels that harm the performance of the quantized models.
    \item We propose Diverse Sample Generation (DSG) scheme, a novel sample generation method for accurate data-free quantization that enhances the diversity of data by combining two practical techniques: \textit{Slack Distribution Alignment} and \textit{Layerwise Sample Enhancement} to solve the homogenization at distribution and sample levels.
    \item We evaluate the proposed method on the large-scale image classification task and consistently obtain significant improvements over various base models and state-of-the-art (SOTA) post-training quantization methods.
\end{enumerate}

\section{Related Work}
\textbf{Data-Driven Quantization.} Quantization is a potent approach to accelerate the inference phase due to its low-bit representations of weights and activations. However, models often suffer an accuracy degeneration after quantization, especially when quantized to ultra-low bit-width. 
Quantization-aware training is proposed to retrain or fine-tune the quantized model with training/validation data to improve the accuracy, as described in earlier work such as~\cite{gupta2015deep, jacob2018quantization}. They often give satisfactory results, but the training process is computationally expensive and time-consuming. More crucially, the original training/validation data are not always accessible, especially on private and secure occasions.

Post-training quantization focuses on obtaining accurate quantized models with small computation and time cost, which has achieved relatively good performance without any fine-tuning or training process. Particularly, \cite{banner2019posttraining} approximates the optimal clipping value analytically and introduces a bit allocation policy and bias-correction to quantize both activations and weights to 4-bit. \cite{choukroun2019low} formalizes the linear quantization task as a minimum mean squared error problem for both weights and activations. \cite{zhao2019improving} exploits channel splitting to avoid containing outliers. \cite{nagel2020down} proposes AdaRound, a better weight-rounding mechanism for post-training quantization that adapts to the data and the task loss. However, these aforementioned methods also require access to limited data for recovering the performance.

\begin{figure*}[t]
\centering
\vspace{-0.5in}
\includegraphics[width=17cm]{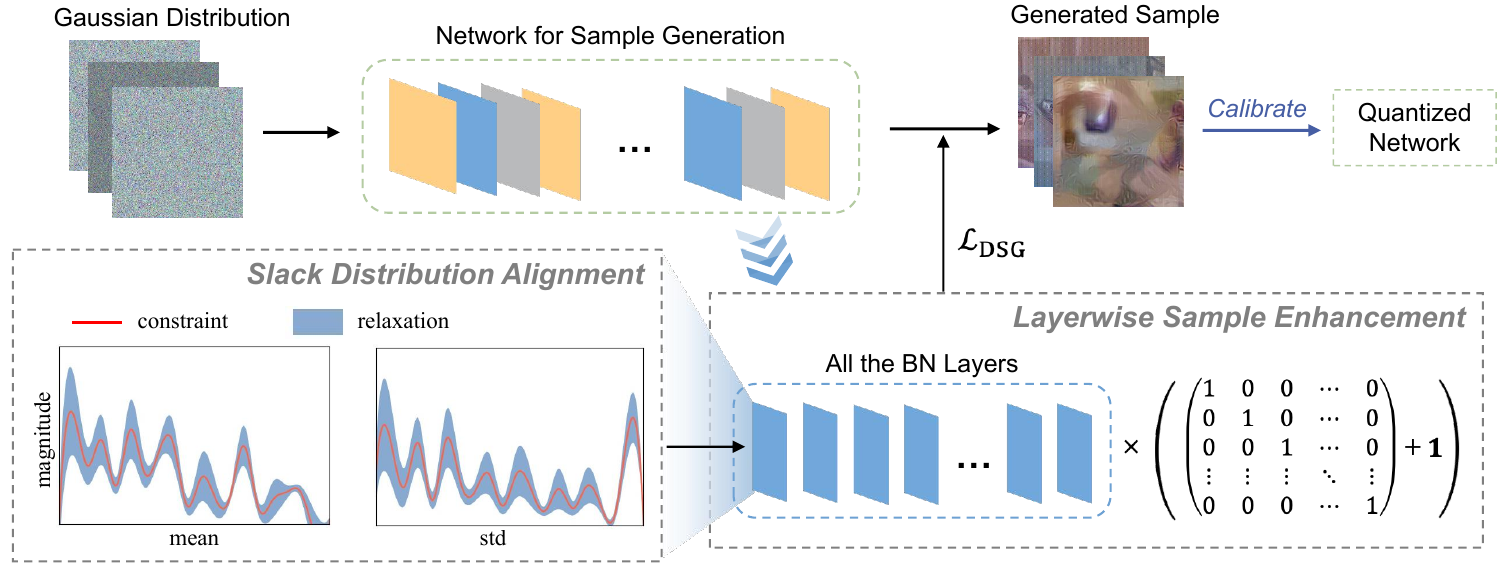}
\vspace{-0.1in}
\caption{The framework of Diverse Sample Generation (DSG) scheme, which consists of Slack Distribution Alignment (SDA) and Layerwise Sample Enhancement (LSE). SDA relaxes BN statistics constraint in each layer, and LSE provides specific loss term for each sample. First, we initialize synthetic data from Gaussian. Then we compute the loss using our SDA and LSE and update synthetic data with this loss. Finally, we calibrate the quantized model with synthetic data.}
\label{fig:structure}
\end{figure*}

\textbf{Data-Free Quantization.} 
Recent work~\cite{Nagel_2019_ICCV, cai2020zeroq, haroush2020knowledge, choi2020datafree, xu2020generative} go further to data-free quantization, which requires neither training nor validation data for quantization. \cite{Nagel_2019_ICCV} uses weight equalization and bias correction to achieve competitive accuracy on layerwise quantization compared with channel-wise quantization. However, it suffers a non-negligible performance drop when quantized to 6 or lower bit-width. While \cite{cai2020zeroq} utilizes mixed-precision quantization with synthetic data to support ultra-low precision quantization. 
\cite{haroush2020knowledge} proposes inception scheme and BN statistics scheme to generate data for calibration and time-consuming knowledge distillation finetuning. Furthermore, \cite{choi2020datafree} proposes a data-free adversarial knowledge distillation method, which minimizes the maximum distance between the outputs of the FP32 teacher model and the quantized student model for any adversarial samples from a generator. 
\cite{xu2020generative} couples the data generation and model finetuning, and uses time-consuming distillation to improve the accuracy of quantized models.
However, \cite{haroush2020knowledge}, \cite{xu2020generative} and \cite{choi2020datafree} have the same limitations as \cite{cai2020zeroq} while generating data that we explain in Section~\ref{sec:homogenization}.

\section{Diverse Sample Generation}
In this section, we revisit the image generation process in data-free quantization and point out the homogenization of the synthetic data in previous work. We present our Diverse Sample Generation (DSG) scheme to diversify the synthetic data for obtaining accurate quantized models.

\subsection{Preliminaries}
Data-free quantization methods are proposed to quantize the FP32 model, and ZeroQ~\cite{cai2020zeroq} is a typical representation of these work, which is proposed to learn an input data distribution that best matches the BN statistics, \ie, the mean and standard deviation, by solving the following optimization problem:
\begin{equation}
\label{eq:BNS}
\min\limits_{\mathbf{x}^s}\mathcal{L}_\textrm{BN}=
\frac{1}{N}\sum\limits_{i=0}^N
\left(\left\|\boldsymbol{\tilde{\mu}}_{i}^s-\boldsymbol{\mu}_{i}\right\|_{2}^{2}+\left\|\boldsymbol{\tilde{\sigma}}_{i}^s-\boldsymbol{\sigma}_{i}\right\|_{2}^{2}\right)
\end{equation}
where $\mathcal{L}_\textrm{BN}$ is the BN statistics loss to be minimized, $\mathbf{x}^s$ is the synthetic input data.
$\boldsymbol{\tilde{\mu}}_{i}^s/\boldsymbol{\tilde{\sigma}}_{i}^s$ are the mean/standard deviation of the feature distribution of synthetic data at the $i$-th BN layer, $\boldsymbol{\mu}_{i}/\boldsymbol{\sigma}_{i}$ are mean/standard deviation parameters stored in $i$-th BN layer of pre-trained FP32 model. 

\subsection{Homogenization of Synthetic Data}
\label{sec:homogenization}
Although many generative data-free quantization methods have been proposed to resolve the problem of accessing real data, they are considered to suffer a huge drop in performance compared with post-training quantization calibrated with real data. We explore the commonly practiced data synthesizing processes and find that the homogenization issue exists at two levels, which degrades the fidelity and quality of synthesized data and thus behave differently with real images:

\noindent\textbf{1) Distribution level homogenization:} 
BN statistics loss in Eq.~(\ref{eq:BNS}) strictly constrains the feature distribution of the synthetic data, aiming to generate samples that exhibit similarities to original training data, which is generally regarded as the upper limit of the performance of synthetic data. 
However, fitting BN statistics obtained from training data is not equivalent to imitating real data. 
As shown in Figure~\ref{fig:distribution_a}, the distribution of synthetic data generated by ZeroQ is almost in the immediate vicinity of the BN statistics. In contrast, the distribution of the real samples deviates from BN statistics.
Therefore, this constraint causes the feature distribution of synthetic data overfitting to the BN statistics in each layer, as so-called distribution homogenization. 


\noindent\textbf{2) Sample level homogenization:} 
Besides homogenization at the distribution level, the application of $\mathcal{L}_\textrm{BN}$ also leads to the homogenization at the sample level.
Existing methods, such as ZeroQ, generate a batch of synthetic data to calibrate or finetune quantized models. 
However, since all the samples are initialized and optimized by the same objective function, the statistics of each sample are quite similar, 
while those of real samples are more versatile. 
As shown in Figure~\ref{fig:distribution_b}, the feature distribution statistics of ZeroQ samples are almost overlapping, while the real-world images have larger variance of statistic distribution. That is to say, ZeroQ data suffers the homogenization at the sample level. Thus, it cannot be applied to determine the clip values of activations for quantized models in place of the real data. Otherwise, it might cause a significant performance drop.

As described above, homogenization exists in the data generation process in previous work of synthetic data-free quantization, including distribution level and sample level, which results in the synthetic data lacking diversity, consequently preventing the quantized models from good performance. In this paper, we propose a novel synthetic data-free quantization method, namely \textbf{D}iverse \textbf{S}ample \textbf{G}eneration (DSG) scheme, which aims to address this issue by enhancing the diversity of synthetic data. 
Calibrating with data generated by our synthesizing scheme, quantized models gain non-negligible improvements in accuracy.

\subsection{Slack Distribution Alignment}
\label{sec:slack}

We propose the Slack Distribution Alignment (SDA) to eliminate the distribution homogenization, which relaxes BN statistics constraint.
Intuitively, we introduce margins for mean and standard deviation statistics of activations, respectively, which allow sufficient distribution variation.
In more detail, we add the relaxation constants to the original BN statistics loss function to tackle the distribution homogenization issue. The loss term of SDA of $i$-th BN layer $l_{\textrm{SDA}_i}$ is defined as follow:
\begin{equation}
\label{eq:SDA}
\begin{aligned}
l_{\textrm{SDA}_i}=&
\left\|\max\left(|\boldsymbol{\tilde{\mu}}_{i}^s-\boldsymbol{\mu}_{i}|-\delta_i, 0\right)\right\|_{2}^{2}\\
+&\left\|\max\left(|\boldsymbol{\tilde{\sigma}}_{i}^s-\boldsymbol{\sigma}_{i}|-\gamma_i, 0\right)\right\|_{2}^{2}
\end{aligned}
\end{equation}
where $\delta_i$ and $\gamma_i$ denote the relaxation constants for the mean and standard deviation statistics of features at the $i$-th BN layers, we admit a gap between the statistics of synthetic data and the statistic parameters of BN layers. Within a specific range, the statistics of synthetic data can fluctuate with relaxed constraints. Thus the feature distribution of synthetic data becomes more diverse, as shown in Figure~\ref{fig:distribution_a}.

A significant challenge is determining the values of $\delta_i$ and $\gamma_i$ without any real data access. Since real data performs well in calibrating the quantized model, the gap between the feature statistics of real data and BN statistics is considered as a reasonable reference, even an optimal degree of relaxation.
Since the neural input is a sum of many inputs, the Gaussian assumption can be seen as a common approximation. From the central limit theorem, we expect it to be approximately Gaussian distribution~\cite{DBLP:journals/corr/abs-1802-05668}.
Therefore, we propose to use synthetic data randomly sampled from Gaussian distribution to determine $\delta_i$ and $\gamma_i$. \textbf{First}, we initialize $1024$ synthetic samples by the Gaussian distribution with $\mu=0$ and $\sigma=1$. We input the synthetic samples into models and save the feature statistics at all BN layers, specifically the mean and standard deviation of the feature distribution. \textbf{Second}, we calculate the gap between the saved statistics and the corresponding BN statistics. We take the percentile of the absolute values of the gaps as $\delta_i$ and $\gamma_i$. Formally, $\delta_i$ and $\gamma_i$ are defined as follows:
\begin{equation}
\label{eq:mn}
\begin{aligned}
\delta_i=\left|\boldsymbol{\tilde\mu}^{0}_{i}-\boldsymbol{\mu}_{i}\right|_{\epsilon} \quad
\gamma_i=\left|\boldsymbol{\tilde\sigma}^{0}_{i}-\boldsymbol{\sigma}_{i}\right|_{\epsilon}
\end{aligned}
\end{equation}
where $\boldsymbol{\tilde\mu}^{0}_{i}/\boldsymbol{\tilde\sigma}^{0}_{i}$ are mean/standard deviation of the feature of Gaussian initialized data $\mathbf{x}^{0}$ at the $i$-th BN layer.
$\left|\boldsymbol{\tilde\mu}^{0}_{i}-\boldsymbol{\mu}_{i}\right|_{\epsilon}$ and $\left|\boldsymbol{\tilde\sigma}^{0}_{i}-\boldsymbol{\sigma}_{i}\right|_{\epsilon}$ denote the $\epsilon$ percentile of $\left|\boldsymbol{\tilde\mu}^{0}_{i}-\boldsymbol{\mu}_{i}\right|$ and $\left|\boldsymbol{\tilde\sigma}^{0}_{i}-\boldsymbol{\sigma}_{i}\right|$, respectively. The value of $\epsilon$ in range $(0, 1]$ determines the values of $\delta_i$ and $\gamma_i$, further settling the degree of relaxation. When the value of $\epsilon$ becomes larger, the constraints in Eq.~(\ref{eq:SDA}) are more relaxing. The default value of $\epsilon$ is set as $0.9$ to prevent the influence of outliers.



\subsection{Layerwise Sample Enhancement}
Existing generative data-free quantization methods always use the same objective function to optimize all samples of data. Specifically, as shown in Eq.~(\ref{eq:BNS}), the loss terms of all layers are equivalently summed. Meanwhile, the optimizing strategy is invariant among samples. In other words, all the samples have the same attention to each layer, which leads to homogeneity at the sample level. 

In order to enhance the diversity at the sample level, we propose \textbf{L}ayerwise \textbf{S}ample \textbf{E}nhancement (LSE), which reinforces the loss of a specific layer for each sample. 
Specifically, the loss function of every synthetic image in a batch may be different.
In fact, for a network with $N$ BN layers, LSE can provide $N$ loss terms and apply each of them to the specific data sample.
Here, we suppose to generate a batch of images, and the batch size is set as $N$, equaling to the number of BN layers of the model. Therefore, the loss term $\mathcal{L}$ of LSE for this batch is defined as:
\begin{equation}
\label{eq:LSE}
\mathcal{L}=\frac{1}{N}\cdot\mathbf{1}^T\left(\left(\mathbf{I}+\mathbf{1}\mathbf{1}^T\right)\mathbf{L}\right)
\end{equation}
where $\mathbf{I}$ is an ${N}$-dimensional identity matrix, $\mathbf{1}$ is an ${N}$ dimension column vector of all ones, $\mathbf{L}=\{l_0, l_1, ..., l_N\}^T$ represents the vector containing loss terms of all BN layers.
We define $\mathbf{X}_\textrm{LSE}=\mathbf{I}+\mathbf{1}\mathbf{1}^T$ as the enhancement matrix, and the Eq.~(\ref{eq:LSE}) can be simplified as:
\begin{equation}
\label{eq:LSE_expand}
\mathcal{L}=\frac{1}{N}\cdot\mathbf{1}^T\left(\mathbf{X}_\textrm{LSE}\mathbf{L}\right)
\end{equation}
where $\mathbf{X}_\textrm{LSE}\mathbf{L}$ can be seen as a $N$-dimensional column vector, the $i$-th element of which represents the loss function of the $i$-th image in this batch. 
Thus, we impose a unique constraint on each sample of this batch and jointly optimize the whole batch.


For a network with ${N}$ BN layers, LSE can simultaneously generate various samples in a batch, and each kind of sample has a enhancement on a particular layer. Specifically, the SDA is applied to obtain the loss term of layers $\textbf{L}_\textrm{SDA}=\{{l}_{\textrm{SDA}_0}, {l}_{\textrm{SDA}_1}, \dots, {l}_{\textrm{SDA}_N}\}^T$, and our DSG loss $\mathcal{L}_\textrm{DSG}$ is defined as:
\begin{equation}
\label{eq:LSE_expand_2}
\begin{aligned}
\mathcal{L}_\textrm{DSG}=\frac{1}{N}\cdot\mathbf{1}^T\left(\mathbf{X}_\textrm{LSE}\mathbf{L}_\textrm{SDA}\right)
\end{aligned}
\end{equation}
Compared with the synthetic data generated with BN statistics loss, images generated by the DSG scheme may have various distributions in one layer. This significantly improves the diversity of synthetic data, which is important to calibrate the quantized models. 
As shown in Figure~\ref{fig.homogenization}, significant fluctuation exists in the distribution statistics of synthetic samples generated by the DSG scheme, which is closer to the behavior of real data and does not strictly fit the BN statistics.

Our DSG scheme applies both SDA and LSE to tackle the homogenization issues at both distribution and sample levels and generate diverse samples for accurate data-free quantization. 
Figure~\ref{fig:structure} shows the whole process of the DSG scheme, and it is summarized in Algorithm~\ref{alg:1}.
Instead of imposing a strict constraint on all samples, we relax this constraint using SDA and introduce LSE to reinforce the loss of a specific layer for one sample to mitigate the homogenization at two levels. Therefore, synthetic data generated by the DSG scheme performs better in quantization than those generated by existing generative data-free quantization methods, even can take the place of real data in SOTA post-training quantization methods when real data is not available. 

\renewcommand{\algorithmicrequire}{\textbf{Input:}}
\renewcommand{\algorithmicensure}{\textbf{Output:}}
\begin{algorithm}[t]
\small
    \caption{The generation process of our DSG scheme.}
    \label{alg:1}
    \KwIn{pretrained model $\textrm{M}$ with $N$ BN layers, training iterations $T$.}
    \KwOut{synthetic data: $\mathbf{x}^s$}
    Initialize $\mathbf{x}^s$ from Gaussian distribution $\mathcal{N}(0,1)$\;
    Initialize $\mathbf{x}^{0}$ from Gaussian distribution $\mathcal{N}(0,1)$\;
    Get $\boldsymbol{\mu}_{i}$ and $\boldsymbol{\sigma}_{i}$ from BN layers of $\textrm{M}$, $i=1,2,\dots, N$\;
    Forward propagate $\textrm{M}(\mathbf{x}^{0})$ and gather activations\;
    Compute $\delta_i$ and $\gamma_i$ based on Eq.~(\ref{eq:mn})\;
    
    \For{all $t=1,2,\dots, T$}
    {
    Forward propagate $\textrm{M}(\mathbf{x}^s)$ and gather activations\;
    Get $\boldsymbol{\tilde{\mu}}_{i}^s$ and $\boldsymbol{\tilde{\sigma}}_{i}^s$ from activations\;
    Compute all ${l_\textrm{SDA}}_i$ based on Eq.~(\ref{eq:SDA})\;
    Compute $\mathcal{L}_\textrm{DSG}$ based on Eq.~(\ref{eq:LSE_expand})\;
    Descend $\mathcal{L}_\textrm{DSG}$ and update $\mathbf{x}^s$\;
    }
    Get synthesized data $\mathbf{x}^s$;
\end{algorithm}

\subsection{Analysis and Discussion}
\begin{figure}[t]
\centering
\includegraphics[width=8.2cm]{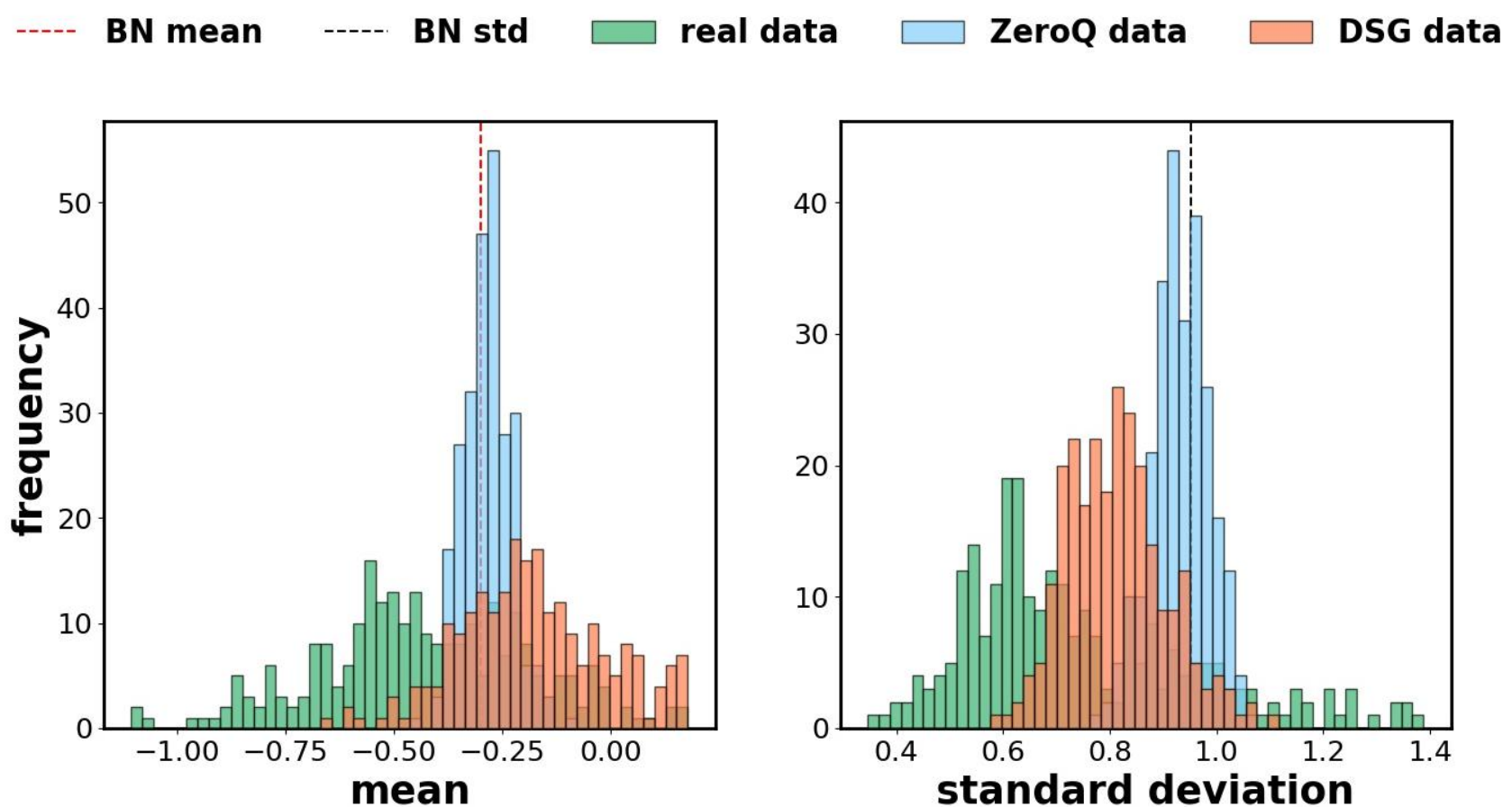}
\vspace{-0.1in}
\caption{Mean and standard deviation of the activations in one channel of ResNet-18 when feeding different types of data (with 256 samples), including real data and synthetic data (generated by ZeroQ and DSG). Each sample generated by ZeroQ behaves similarly overfitting BN statistics compared with real data, which shows the homogenization at both distribution and sample levels. Our DSG data shows the diversity close to real data, which is the key to obtaining the accurate quantized model.}
\label{fig.homogenization}
\end{figure}

We further present discussion with visualization results for our DSG scheme.
The distribution of statistics is shown in Figure~\ref{fig.homogenization}, including that of the real data, ZeroQ data, and DSG data. The figure illustrates the homogenization at the aforementioned two levels.

As shown in Figure~\ref{fig.homogenization}, there is a significant offset between feature statistics of DSG samples and the BN statistics, which is similar to the behavior of real data samples.
While the feature statistics of ZeroQ synthetic samples almost obeys the BN statistics.
This phenomenon proves that our DSG scheme diversifies the synthetic data at the distribution level. Specifically, the SDA slacks the constraint of statistics during the generation process to make the distribution diverse.

Moreover, with a larger variance of the statistic distribution on both mean and standard deviation, the statistics of DSG synthetic data are more disperse, which behave almost in the same as real data. On the contrary, statistics of ZeroQ data seem to be centralized. This phenomenon results from both SDA and LSE, which jointly enhance the data diversity at the sample level. The results imply that since the statistics are widely dispersed, the synthetic data might be more plentiful in feature and have more comprehensive content.

\section{Experiments}

In this section, we conduct experiments on the benchmark ImageNet dataset (ILSVRC12)~\cite{Deng2009ImageNet} large-scale image classification tasks to verify the effectiveness of the proposed DSG scheme and compare it with other state-of-the-art (SOTA) data-free quantization methods.

\textbf{DSG scheme:} Our DSG scheme is implemented by PyTorch for its high flexibility and powerful automatic differentiation mechanism. The proposed DSG scheme is used for generating synthetic data for calibration, and the effectiveness is evaluated by measuring the accuracy of quantized models with various quantization methods, such as Percentile, EMA, and MSE.
Also, we further apply the synthetic data generated by the DSG scheme to the state-of-the-art post-training quantization method: AdaRound~\cite{nagel2020down}.

\textbf{Network Architectures:} To prove the versatility of our DSG scheme, we employ various widely-used network architectures, including ResNet-18, ResNet-50, SqueezeNext, InceptionV3, and ShuffleNet. We also evaluate our DSG scheme with various bit-widths, including W4A4 (means 4-bit Weight and 4-bit Activation), W6A6, W8A8, \etc.

\textbf{Initialization:} For fair comparison, when evaluating our DSG scheme on various network architectures, we mostly follow the hyper-parameter settings (\eg, the number of iterations to generate synthetic data) of their original papers or released official implementations. The data generated by our DSG scheme is initialized by Gaussian random initialization. We adopt Adam~\cite{kingma2014adam} as optimization algorithm in our experiments.

\begin{table}[htb]
    \caption{Ablation study for DSG scheme on ResNet-18. We abbreviate quantization bits used for weights as “W-bit” (for activations as “A-bit”), top-1 test accuracy as “Top-1.” }
    \vspace{-0.14in}
    \label{ablation_exp}
	\centering
    \setlength{\tabcolsep}{2.0mm}
    {\small
    \begin{tabular}{lccc}
		\toprule
		{Method}  &{W-bit}  &{A-bit}  &{Top-1} \\
		\midrule
		Baseline  &32 &32 &71.47  \\
		\midrule
		Vanilla (ZeroQ) &4  &4  &26.04  \\
		Layerwise Sample Enhancement &4  &4  &27.12  \\
		Slack Distribution Alignment    &4  &4  &33.39  \\
		{DSG} (Ours)    &4  &4& \textbf{34.53}  \\
		\bottomrule
	\end{tabular}
	}
\end{table}

\subsection{Ablation Study}



We perform ablation studies to investigate the effect of components of the proposed DSG scheme, with the ResNet-18 model on the ImageNet dataset. 
We evaluate our method on W4A4 bit-width, which can reveal the effect of each part most obviously. 

\subsubsection{Effect of SDA}
\begin{figure}[t]
\centering
\vspace{-0.2in}
\includegraphics[width=6.8cm]{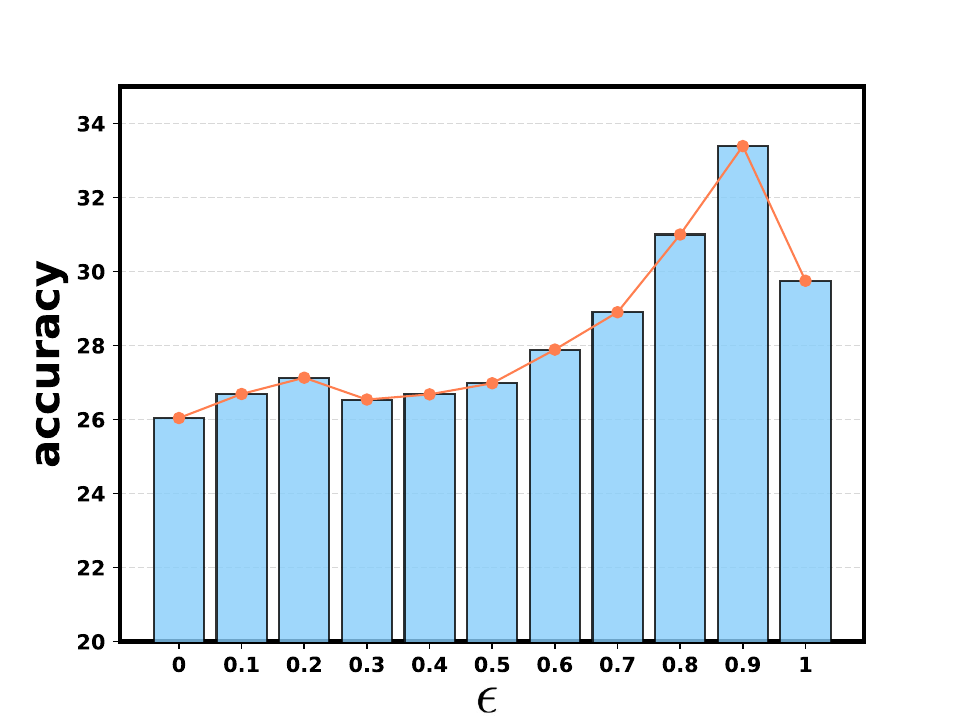}
\vspace{-0.1in}
\caption{The accuracy comparison of different $\epsilon$ values in Eq.~(\ref{eq:mn}) on ResNet-18. 
As $\epsilon$ varies from $0$ to $0.9$, the final accuracy is mainly on the rise.
But it suffers a significant drop caused by the outliers when $\epsilon=1$.}
\label{fig:Eqmn.perc}
\end{figure}

We first evaluate the effectiveness of SDA. As described in Section~\ref{sec:slack}, the degree of the slack is adaptively determined by the value of $\epsilon$ in SDA. Thus, in Figure~\ref{fig:Eqmn.perc}, we explore the impact of different values of $\epsilon$ in Eq.~(\ref{eq:mn}).
Since $\epsilon$ is in the range of $(0, 1]$, we adopt a moderate interval, which is $0.1$. Furthermore, we add $\epsilon=0$ to complement the vanilla case that the constraints are not slacked.

Table~\ref{ablation_exp} shows that if SDA is absent, the performance of the quantized network drops significantly by 7.41\% compared with that using two methods as a whole, which shows that SDA is essential and even provides the major contribution to the final performance.
In more detail, as shown in Figure~\ref{fig:Eqmn.perc}, when $\epsilon$ increases from $0$ to $0.9$, the final performance of the quantized model increase steadily. However, when $\epsilon$ is set to $1$, the accuracy suffers a huge drop. The phenomenon forcefully confirms that the distribution slack introduced by our SDA relaxes the constraints and allows the statistics of synthetic data to have a certain degree of offset, which enhances the diversity of features and consequently boosts the performance of the quantized model. 
Nevertheless, suppose $\epsilon$ is set to $1$, which means that all the outliers in $\boldsymbol{\tilde\mu}^{0}_{i}$ and $\boldsymbol{\tilde\sigma}^{0}_{i}$ are taken into consideration, the degree of slack becomes out of bounds and the feature distribution of synthetic data might be far away from the reasonable range. Therefore, the default value of $\epsilon$ is set to $0.9$ empirically to mitigate the impact of outliers.

\subsubsection{Effect of LSE}

Table~\ref{ablation_exp} also shows that LSE contributes to the final performance. From the results, LSE achieves up to 1.08\% improvement compared with ZeroQ, which is slight but non-negligible.

Experiment results demonstrate that these two methods are compatible and jointly boost the performance of the quantized model.
From the results, the improvements brought by these two parts are superimposed because they focus on different root causes, and the processes do not interfere with each other. Consequently, quantized models calibrated with samples generated by our methods surpass vanilla competitors by 8.49\%.
In short, SDA prevents the synthetic samples from overfitting to BN statistics, and LSE makes the samples have different focuses on the statistics of specific layers. 

\begin{table}[htb]
    \caption{Quantization results of ResNet-18 and ResNet-50 on ImageNet. Here, “No D” means that none of the data is used to assist quantization, “No FT” stands for no fine-tuning (retraining). “Real Data” represents using real training data and quantization methods in ZeroQ (without any fine-tuning). Our DSG outperforms the SOTA data-free quantization methods and quantization methods requiring real data, such as ZeroQ, DFQ, DFC, Integer-Only~\cite{jacob2018quantization}, and OCS~\cite{zhao2019improving}.
    }
    \vspace{-0.2in}
    \subtable[ResNet-18]{
    \label{tb:exp_res18}
	\centering
	\setlength{\tabcolsep}{2.1mm}
    {\small
    \begin{tabular}{lccccc}
		\toprule
		{Method} &{No D}  &{No FT}  &{W-bit}  &{A-bit}  &\tabincell{c}{{Top-1}}\\
		\midrule
		Baseline   &--  &-- &32 &32 &71.47  \\
		\midrule
		Real Data  &{\footnotesize\XSolidBrush}   &{\footnotesize\Checkmark} &4  &4  &31.86  \\
		\midrule
		ZeroQ  &{\footnotesize\Checkmark}   &{\footnotesize\Checkmark} &4  &4  &26.04  \\
		{DSG} (Ours)  &{\footnotesize\Checkmark}   &{\footnotesize\Checkmark}  &4  &4 &\textbf{34.53}  \\
		\midrule
		Real Data  &{\footnotesize\XSolidBrush}   &{\footnotesize\Checkmark} &6  &6  &70.62  \\
		\midrule
		Integer-Only  &{\footnotesize\XSolidBrush}   &{\footnotesize\XSolidBrush} &6  &6  &67.30  \\
		DFQ  &{\footnotesize\Checkmark}   &{\footnotesize\Checkmark} &6  &6  &66.30  \\
		ZeroQ  &{\footnotesize\Checkmark}   &{\footnotesize\Checkmark} &6  &6  &69.74  \\
		{DSG} (Ours)  &{\footnotesize\Checkmark}   &{\footnotesize\Checkmark}  &6  &6 &\textbf{70.46}  \\
		\midrule
		Real Data  &{\footnotesize\XSolidBrush}   &{\footnotesize\Checkmark} &8  &8  &71.44  \\%
		\midrule
		RVQuant  &{\footnotesize\XSolidBrush}   &{\footnotesize\XSolidBrush} &8  &8  &70.01  \\
		DFQ  &{\footnotesize\Checkmark}   &{\footnotesize\Checkmark} &8  &8  &69.70  \\
		DFC  &{\footnotesize\Checkmark}   &{\footnotesize\XSolidBrush} &8  &8  &69.57  \\
		ZeroQ  &{\footnotesize\Checkmark}   &{\footnotesize\XSolidBrush} &8  &8  &71.43  \\
		{DSG} (Ours)  &{\footnotesize\Checkmark}   &{\footnotesize\Checkmark}  &8  &8 &\textbf{71.49}  \\
		\bottomrule
	\end{tabular}
	}
	}
    \subtable[ResNet-50]{
    \label{tb:exp_res50}
        \centering
        {\small
        \begin{tabular}{lccccc}
    		\toprule
    		{Method} &{No D}  &{No FT}  &{W-bit}  &{A-bit}  &\tabincell{c}{{Top-1}}\\
    		\midrule
    		Baseline   &--  &-- &32 &32 & 77.72 \\%
    		\midrule
    		Real Data  &{\footnotesize\XSolidBrush}   &{\footnotesize\Checkmark} &6  &6  &75.52  \\%
    		\midrule
    		OCS  &{\footnotesize\XSolidBrush}   &{\footnotesize\Checkmark} &6  &6  & 74.80 \\%
    		ZeroQ  &{\footnotesize\Checkmark}   &{\footnotesize\Checkmark} &6  &6  & 75.56 \\%
    		{DSG} (Ours)  &{\footnotesize\Checkmark}   &{\footnotesize\Checkmark} &6  &6  & \textbf{76.07} \\%
    		\midrule
    		Real Data  &{\footnotesize\XSolidBrush}   &{\footnotesize\Checkmark} &8  &8  &77.70  \\%
    		\midrule
    		ZeroQ  &{\footnotesize\Checkmark}   &{\footnotesize\Checkmark} &8  &8  & 77.67 \\%
    		{DSG} (Ours)  &{\footnotesize\Checkmark}   &{\footnotesize\Checkmark} &8  &8  &  \textbf{77.68}\\%
    		\bottomrule
    	\end{tabular}
    	}
    }
    \vspace{-0.1in}
\end{table}

\begin{table}[htbp]
    \centering
    \vspace{-0.2in}
    \caption{Quantization results of SqueezeNext, InceptionV3 and ShuffleNet on ImageNet.}
    \subtable[SqueezeNext]{
        \label{tb:exp_sqxt}
        \centering
        {\small
        \begin{tabular}{lccccc}
    		\toprule
    		{Method} &{No D}  &{No FT}  &{W-bit}  &{A-bit}  &\tabincell{c}{{Top-1}}\\
    		\midrule
    		Baseline   &--  &-- &32 &32 &  69.38\\%
    		\midrule
    		Real Data  &{\footnotesize\XSolidBrush}   &{\footnotesize\Checkmark} &6  &6  &62.88  \\%
    		\midrule
    		ZeroQ  &{\footnotesize\Checkmark}   &{\footnotesize\Checkmark} &6  &6  &39.83  \\%
    		{DSG} (Ours)  &{\footnotesize\Checkmark}   &{\footnotesize\Checkmark} &6  &6  &\textbf{60.50}  \\%
    		\midrule
    		Real Data  &{\footnotesize\XSolidBrush}   &{\footnotesize\Checkmark} &8  &8  &69.23  \\%
    		\midrule
    		ZeroQ  &{\footnotesize\Checkmark}   &{\footnotesize\Checkmark} &8  &8  &68.01  \\%
    		{DSG} (Ours)  &{\footnotesize\Checkmark}   &{\footnotesize\Checkmark} &8  &8  &\textbf{69.27}  \\%
    		\bottomrule
    	\end{tabular}
    	}
    }
    \subtable[InceptionV3]{
        \label{tb:exp_incepv3}
        \centering
        {\small
        \begin{tabular}{lccccc}
    		\toprule
    		{Method} &{No D}  &{No FT}  &{W-bit}  &{A-bit}  &\tabincell{c}{{Top-1}}\\
    		\midrule
    		Baseline   &--  &-- &32 &32 &  78.80\\%
    		\midrule
    		Real Data  &{\footnotesize\XSolidBrush}   &{\footnotesize\Checkmark} &4  &4  & 23.23 \\%
    		\midrule
    		ZeroQ  &{\footnotesize\Checkmark}   &{\footnotesize\Checkmark} &4  &4  & 12.00 \\%
    		{DSG} (Ours)  &{\footnotesize\Checkmark}   &{\footnotesize\Checkmark} &4  &4  & \textbf{34.89} \\%
    		\midrule
    		Real Data  &{\footnotesize\XSolidBrush}   &{\footnotesize\Checkmark} &6  &6  & 77.96  \\%
    		\midrule
    		ZeroQ  &{\footnotesize\Checkmark}   &{\footnotesize\Checkmark} &6  &6  & 75.14 \\%
    		{DSG} (Ours)  &{\footnotesize\Checkmark}   &{\footnotesize\Checkmark} &6  &6  & \textbf{76.52} \\%
    		\midrule
    		Real Data  &{\footnotesize\XSolidBrush}   &{\footnotesize\Checkmark} &8  &8  & 78.78  \\%
    		\midrule
    		ZeroQ  &{\footnotesize\Checkmark}   &{\footnotesize\Checkmark} &8  &8  & 78.70 \\%
    		{DSG} (Ours)  &{\footnotesize\Checkmark}   &{\footnotesize\Checkmark} &8  &8  & \textbf{78.81}  \\%
    		\bottomrule
    	\end{tabular}
    	}
    }
    \subtable[ShuffleNet]{
        \label{tb:exp_shuffle}
        \centering
        {\small
        \begin{tabular}{lccccc}
    		\toprule
    		{Method} &{No D}  &{No FT}  &{W-bit}  &{A-bit}  &\tabincell{c}{{Top-1}}\\
    		\midrule
    		Baseline   &--  &-- &32 &32 & 65.07 \\%
    		\midrule
    		Real Data  &{\footnotesize\XSolidBrush}   &{\footnotesize\Checkmark} &6  &6  & 44.75 \\%
    		\midrule
    		ZeroQ  &{\footnotesize\Checkmark}   &{\footnotesize\Checkmark} &6  &6  & 39.92\\%
    		{DSG} (Ours)  &{\footnotesize\Checkmark}   &{\footnotesize\Checkmark} &6  &6  &  \textbf{44.88}\\%
    		\midrule
    		Real Data  &{\footnotesize\XSolidBrush}   &{\footnotesize\Checkmark} &8  &8  & 64.52 \\%
    		\midrule
    		ZeroQ  &{\footnotesize\Checkmark}   &{\footnotesize\Checkmark} &8  &8  & 64.46 \\%
    		{DSG} (Ours)  &{\footnotesize\Checkmark}   &{\footnotesize\Checkmark} &8  &8  & \textbf{64.77} \\%
    		\bottomrule
    	\end{tabular}
    	}
    }
    \vspace{-0.1in}
\end{table}

\subsection{Comparison with SOTA methods}
We evaluate our DSG scheme on ImageNet dataset for large-scale image classification task, and analysis the performance over various network architectures, including ResNet-18 (Table~\ref{tb:exp_res18}), ResNet-50 (Table~\ref{tb:exp_res50}), SqueezeNext (Table~\ref{tb:exp_sqxt}), InceptionV3 (Table~\ref{tb:exp_incepv3}), and ShuffleNet (Table~\ref{tb:exp_shuffle}). As mentioned above, we adopt SDA for all samples and apply LSE to part of them while generating synthetic data. The bit-width of the quantized model is marked as W$w$A$a$ standing for $w$-bit weight and $a$-bit activation, which is set to W8A8, W6A6, W4A4, \etc.

\begin{table}[htb]
    \caption{Uniform post-quantization on ImageNet with ResNet-18. We evaluation our DSG scheme on various post-training quantization methods (Percentile, EMA, MSE), and Vanilla means the quantization method adopted by ZeroQ, which simply obtains the quantizer by the maximum and minimum of the weight and activation.}
    \vspace{-0.25in}
    \label{tb:exp_quantization_method}
	\centering
    \setlength{\tabcolsep}{1.8mm}
    {\small
    \begin{tabular}{lccccc}
		\toprule
		{Method} &{No D}  &{W-bit}  &{A-bit} & {Quant}  &\tabincell{c}{{Top-1}}\\
		\midrule
		Baseline   &--  &-- &32 &32 &71.47  \\
		\midrule
		Real Data  &{\footnotesize\XSolidBrush}   &4  &4  & Vanilla &31.86  \\
		\midrule
		ZeroQ  &{\footnotesize\Checkmark}   &4  &4 & Vanilla  &26.04  \\
		{DSG} (Ours)  &{\footnotesize\Checkmark}   &4  &4 & Vanilla &\textbf{34.53}  \\
		\midrule
		Real Data  &{\footnotesize\XSolidBrush}   &4  &4  & Percentile &42.83  \\
		\midrule
		ZeroQ  &{\footnotesize\Checkmark}  &4  &4 &Percentile &32.24  \\
		{DSG} (Ours)  &{\footnotesize\Checkmark}  &4  &4 &Percentile &\textbf{38.76}  \\
		\midrule
		Real Data  &{\footnotesize\XSolidBrush}  &4  &4  & EMA & 42.67  \\
		\midrule
		ZeroQ  &{\footnotesize\Checkmark}  &4  &4  &EMA &32.31  \\
		{DSG} (Ours)  &{\footnotesize\Checkmark}  &4  &4 &EMA &\textbf{35.18}  \\
		\midrule
		Real Data  &{\footnotesize\XSolidBrush}  &4  &4  &MSE &41.45  \\
		\midrule
		ZeroQ  &{\footnotesize\Checkmark}   &4  &4  &MSE &39.39  \\
		{DSG} (Ours)   &{\footnotesize\Checkmark}  &4  &4 &MSE  &\textbf{40.00}  \\
		\bottomrule
	\end{tabular}
	}
	\vspace{-0.1in}
\end{table}

Our method outperforms previous state-of-the-art methods in various bit-width settings and is even better than those requiring real data to calibrate quantized models directly. 
In W8A8 cases, our method surpasses previous post-training quantization and quantization-aware training methods, such as DFC~\cite{haroush2020knowledge}, DFQ~\cite{Nagel_2019_ICCV}, and RVQuant~\cite{park2018valueaware}, as shown in the bottom part in Table~\ref{tb:exp_res18}. And meanwhile, we consistently observe significant improvement over various network architectures, as shown in Table~\ref{tb:exp_res50}, Table~\ref{tb:exp_sqxt}, Table~\ref{tb:exp_incepv3}, and Table~\ref{tb:exp_shuffle}. For instance, our method outperforms ZeroQ on SqueezeNext by 1.26\% on W8A8. 
However, advancement gets more apparent when it goes further to lower bit-width cases. On W6A6, our method significantly surpasses ZeroQ by more than 20\% on SqueezeNext, and even outperforms real data on ShuffleNet by a slight 0.13\%.  

Moreover, we highlight that our DSG achieves significant improvement with even lower bit-width, \ie W4A4 cases. As shown in Table~\ref{tb:exp_res18}, the performance of DSG on ResNet-18 is up to 34.53\%, which is 8.49\% higher than that of ZeroQ, 2.67\% higher than that directly using real data for calibration. As well as on InceptionV3, our method achieves 34.89\% accuracy surpassing ZeroQ and real data by 22.89\% and 11.66\%, respectively, which is quite a wide margin. 

In short, sufficient experiments over various network architectures and different bit-widths demonstrate that the synthetic data generated by the proposed DSG scheme can significantly improve the performance of the quantized model. The results also suggest that the diversity of synthetic data is important for improving the quantized model. Especially when the model is quantized to lower bit-width, the advantages of diversity becomes even more obvious on the final performance. 

We further evaluate the DSG scheme over various quantization methods on W4A4 to verify its versatility. Specifically, we implement Percentile, EMA, and MSE in conjunction with different data generation method. Table~\ref{tb:exp_quantization_method} shows that the synthetic data generated by the DSG scheme achieves leading accuracy regardless of specific quantization methods and surpasses ZeroQ while using Percentile, EMA, and MSE by 6.52\%, 2.87\%, and 0.61\% respectively. The results also demonstrate that DSG scheme is robust and versatile to various conditions. 

\subsection{Evaluation with AdaRound}
Besides the above-mentioned post-training quantization methods aiming to obtain optimal clipping values for activations, we further evaluate our DSG scheme with AdaRound~\cite{nagel2020down}, the novel post-training quantization method which proposes a rounding procedure for quantizing weights. We also conduct experiments in conjunction with other data generation methods. Generating data with labels~\cite{9156818} can extract the class information from the parameters of networks. And image prior~\cite{yin2020dreaming} helps to steer synthetic data away from unrealistic images with no discernible visual information. More specifically, since AdaRound only aims to quantize weights, we preserve full precision for activations in the experiments. And we generate 1024 samples for AdaRound to learn the rounding scheme. The results are shown in Table~\ref{tb:ablation_exp}, and our DSG scheme surpasses ZeroQ under all settings, especially when we quantize the weights to ultra-low bit-width (\ie 3-bit). Combined with labels and image prior approach, our DSG scheme still maintains high accuracy. 

\begin{table}[t]
    \caption{AdaRound on ImageNet with ResNet-18. We evaluate the DSG scheme on AdaRound, one of the SOTA methods of post-training quantization, which learns how to quantize weights using several batches of unlabeled samples. We adopt "Label"~\cite{9156818} and "Image Prior"~\cite{yin2020dreaming} to evaluating our DSG scheme further.}
    \vspace{-0.25in}
    \label{tb:ablation_exp}
	\centering
    \setlength{\tabcolsep}{1.25mm}
    {\small
    \begin{tabular}{lcccccc}
		\toprule
		{Method} &{No D}  &{Label} &{Image Prior}  &{W-bit}  &{A-bit}  &\tabincell{c}{{Top-1}}\\
		\midrule
		Real Data  &{\footnotesize\XSolidBrush} &{\footnotesize\XSolidBrush} 
		&{\footnotesize\XSolidBrush} &3  &32  & 64.16\\
		\midrule
		ZeroQ  &{\footnotesize\Checkmark} &{\footnotesize\XSolidBrush}  &{\footnotesize\XSolidBrush} &3  &32  & 49.86 \\
		DSG (Ours)  &{\footnotesize\Checkmark} &{\footnotesize\XSolidBrush}  &{\footnotesize\XSolidBrush} &3  &32  & \textbf{56.09} \\
		DSG (Ours)  &{\footnotesize\Checkmark} &{\footnotesize\Checkmark}  &{\footnotesize\XSolidBrush} &3  &32  & \textbf{58.27} \\
		DSG (Ours)  &{\footnotesize\Checkmark} &{\footnotesize\Checkmark}  &{\footnotesize\Checkmark} &3  &32  & \textbf{61.32} \\
		\midrule
		Real Data  &{\footnotesize\XSolidBrush} &{\footnotesize\XSolidBrush} 
		&{\footnotesize\XSolidBrush} &4  &32  & 68.42 \\
		\midrule
		ZeroQ  &{\footnotesize\Checkmark} &{\footnotesize\XSolidBrush}  &{\footnotesize\XSolidBrush} &4  &32  & {63.86} \\
		DSG (Ours)  &{\footnotesize\Checkmark} &{\footnotesize\XSolidBrush}  &{\footnotesize\XSolidBrush} &4  &32  & \textbf{66.87} \\
		DSG (Ours)  &{\footnotesize\Checkmark} &{\footnotesize\Checkmark}  &{\footnotesize\XSolidBrush} &4  &32  & \textbf{67.09} \\
		DSG (Ours)  &{\footnotesize\Checkmark} &{\footnotesize\Checkmark}  &{\footnotesize\Checkmark} &4  &32  & \textbf{67.78} \\
		\midrule
		Real Data  &{\footnotesize\XSolidBrush} &{\footnotesize\XSolidBrush} 
		&{\footnotesize\XSolidBrush} &5  &32  & 69.21\\
		\midrule
		ZeroQ  &{\footnotesize\Checkmark} &{\footnotesize\XSolidBrush}  &{\footnotesize\XSolidBrush} &5  &32  & 68.39 \\
		DSG (Ours)  &{\footnotesize\Checkmark} &{\footnotesize\XSolidBrush}  &{\footnotesize\XSolidBrush} &5  &32  & \textbf{68.97} \\
		DSG (Ours)  &{\footnotesize\Checkmark} &{\footnotesize\Checkmark}  &{\footnotesize\XSolidBrush} &5  &32  & \textbf{69.02} \\
		DSG (Ours)  &{\footnotesize\Checkmark} &{\footnotesize\Checkmark}  &{\footnotesize\Checkmark} &5  &32  & \textbf{69.16} \\
		\bottomrule
	\end{tabular}
	}
	\vspace{-0.1in}
\end{table}

\section{Conclusion}
We first revisit the data generation process in data-free quantization and demonstrate that homogenization exists at both distribution and sample level in the data generated by the previous data-free quantization method, which harms the accuracy of quantized models. 
Toward this end, we have represented a novel sample generation method for accurate data-free quantization, dubbed as Diverse Sample Generation (DSG) scheme, to mitigate the homogenization issue.
The proposed DSG scheme consists of Slack Distribution Alignment (SDA) and Layerwise Sample Enhancement (LSE), which are tailored to the aforementioned two levels of homogenization respectively and jointly enhance the diversity of generated data. 
Extensive experiments on multiple network architectures and post-training quantization methods demonstrate the leading accuracy and versatility of the DSG scheme.
Especially on ultra-low bit-width, such as W4A4, our method achieves even up to 22\% improvement.

\noindent \ 

\noindent\textbf{Acknowledgement} This work was supported by National Natural Science Foundation of China (62022009, 61872021), Beijing Nova Program of Science and Technology (Z191100001119050), State Key Lab of Software Development Environment (SKLSDE-2020ZX-06), and SenseTime Research Fund for Young Scholars.

{
\bibliography{egbib}
\bibliographystyle{ieee_fullname}
}

\end{document}



\title{Diversifying Sample Generation for Accurate Data-Free Quantization\\Supplemental Material}
\author{
\fontsize{11.0pt}
{\baselineskip}\selectfont Xiangguo Zhang\thanks{Equal contribution.} \textsuperscript{ 1}, Haotong Qin\textsuperscript{* 1}, Yifu Ding\textsuperscript{1}, Ruihao Gong\textsuperscript{3, 4},\\
Qinghua Yan\textsuperscript{1}, Renshuai Tao\textsuperscript{1}, Yuhang Li\textsuperscript{2}, Fengwei Yu\textsuperscript{3, 4}, Xianglong Liu\textsuperscript{1\thanks{Corresponding author}}\\
\textsuperscript{1}{\fontsize{11.0pt}{\baselineskip}\selectfont Beihang University}\quad
\textsuperscript{2}{\fontsize{11.0pt}{\baselineskip}\selectfont Yale University}\quad
\textsuperscript{3}{\fontsize{11.0pt}{\baselineskip}\selectfont SenseTime Research}\quad
\textsuperscript{4}{\fontsize{11.0pt}{\baselineskip}\selectfont Shanghai AI Laboratory}\\
{\fontsize{8.5pt}{\baselineskip}\selectfont \tt \{xiangguozhang, zjdyf, yanqh, rstao\}@buaa.edu.cn, yuhang.li@yale.edu,}\\
{\fontsize{8.5pt}{\baselineskip}\selectfont \tt \{qinhaotong, xlliu\}@nlsde.buaa.edu.cn, \{gongruihao, yufengwei\}@sensetime.com}
}
\maketitle

\pagestyle{empty}
\thispagestyle{empty}

\section{Visualizations of Activation Statistics from Real and Synthetic Data}

\begin{figure}[t]
\centering
\vspace{-0.3in}
\subfigure[]{
\includegraphics[width=0.49\linewidth]{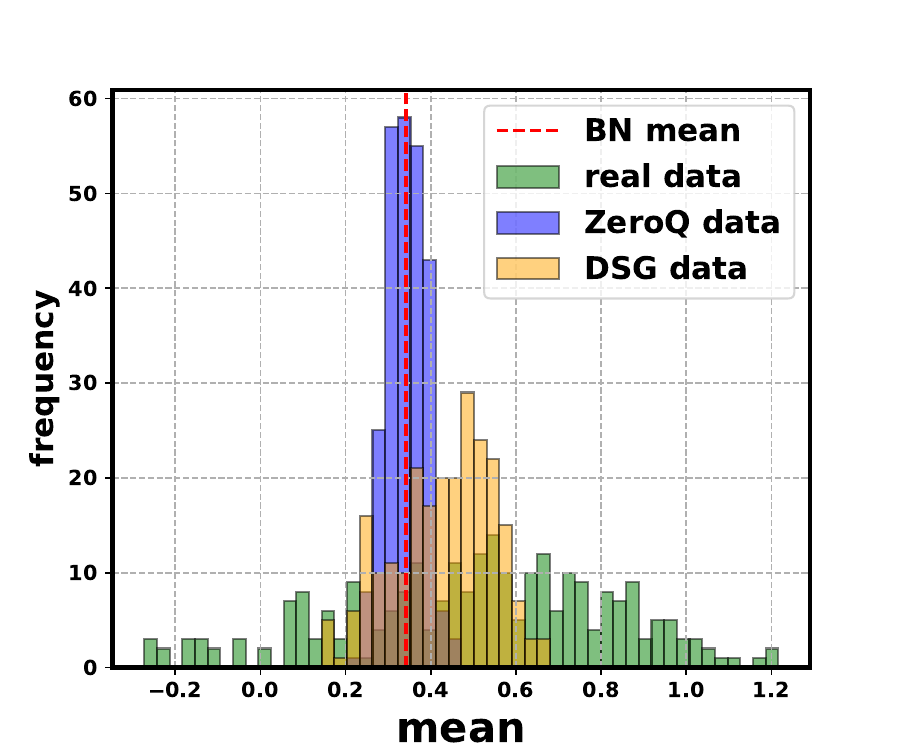}
\label{fig:distribution_a}
}
\vspace{-0.15in}
\hspace{-0.6cm}
\subfigure[]{
\includegraphics[width=0.49\linewidth]{11std (1).pdf}
\label{fig:distribution_b}
}
\vspace{-0.15in}
\subfigure[]{
\includegraphics[width=0.49\linewidth]{15mean (1).pdf}
\label{fig:distribution_c}
}
\hspace{-0.6cm}
\subfigure[]{
\includegraphics[width=0.49\linewidth]{13std (3).pdf}
\label{fig:distribution_d}
}
\vspace{-0.05in}
\caption{Mean and standard deviation of the activations in two different channels of ResNet-18 when feeding different types of data (with 256 samples). (a) and (b) show the mean and deviation of one specific channel, while (c) and (d) show another. 
}
\label{fig:hist}

\end{figure}

In our paper, we have presented the distributions of mean and standard deviation values in one channel (Figure 3 in our main paper). To give ampler evidence, we provide more visualizations for other channels in Figure~\ref{fig:hist}. 
Take Figure~\ref{fig:distribution_a} and \ref{fig:distribution_b} for example, each of them represents the mean or standard deviation in one channel for different types of data. Compared with the statistics of real data which are considered as reasonable references, the mean values of DSG data are more dispersed than those of ZeroQ data, as well as the standard deviation values. Meanwhile, the distributions of ZeroQ in each bar graph are always in the immediate vicinity of BN statistics (the red dash line). Figure~\ref{fig:distribution_c} and \ref{fig:distribution_d} show the case of another channel, which has just the same phenomenon as described above.

To further demonstrate that to what extent our method really affects, we additionally showcase some box figures in the following for auxiliary instruction (see Figure~\ref{fig:box}).
As can be obviously seen from the figure, both the mean and standard deviation statistics of ZeroQ~\cite{cai2020zeroq} data are centralized, which have shorter boxes in Figure~\ref{fig:box_c} and~\ref{fig:box_d}. However, real data (Figure~\ref{fig:box_a} and ~\ref{fig:box_b}) and DSG data (Figure~\ref{fig:box_e} and ~\ref{fig:box_f}) have longer boxes, which implies that the distributions of each sample are more dispersed. 

Take a closer look at those turning points on the line. Each turning point represents the value of BN mean/standard deviation in one layer. The offset between the turning point and the middle of the box at the same column is much smaller in ZeroQ cases, which means the statistics of ZeroQ data mostly fit the distribution of BN statistics. Whereas as for DSG data, the offset is big, beneath or over the middle of the corresponding box, which implies that the statistics of synthetic data are no longer overfitting to BN statistics. 

In short, compared with ZeroQ, distribution statistics of DSG data are much closer to those of real data for two reasons: the dispersion in one layer and the offset to BN statistics. We attribute that to the approaches proposed in our paper, \ie SDA and LSE. The former slacks the alignment of the feature statistics to overcome the overfitting issue and the latter applies the layerwise enhancement to reinforce specific layers. We combine these two approaches and obtain diversified samples.

\begin{figure*}[ht]
\centering
\vspace{-0.5cm}
\subfigure[]{
\includegraphics[width=0.32\linewidth]{image/t_m (2).pdf}
\label{fig:box_a}
}
\vspace{-0.4cm}
\hspace{-0.8cm}
\subfigure[]{
\includegraphics[width=0.32\linewidth]{image/f_m (2).pdf}
\label{fig:box_c}
}
\hspace{-0.8cm}
\subfigure[]{
\includegraphics[width=0.32\linewidth]{image/o_m (2).pdf}
\label{fig:box_e}
}
\subfigure[]{
\includegraphics[width=0.32\linewidth]{image/t_std (2).pdf}
\label{fig:box_b}
}
\hspace{-0.8cm}
\subfigure[]{
\includegraphics[width=0.32\linewidth]{image/f_std (2).pdf}
\label{fig:box_d}
}
\hspace{-0.8cm}
\subfigure[]{
\includegraphics[width=0.32\linewidth]{image/o_std (2).pdf}
\label{fig:box_f}
}
\vspace{-0.4cm}
\caption{Comparison between real data and synthetic data (generated by DSG and ZeroQ) with 256 samples of each. 
}
\label{fig:box}
\end{figure*}

\section{Additional Experiments on Other Dataset and Quantization Methods}

In our paper, we have conducted a bunch of experiments to evaluate the effect of our method in both mitigating the homogenization issue and improving the final performance. To further demonstrate the robustness and the general applicability of our method, we provide additional experiments as corroborations to support our viewpoint.

\begin{table}[b]
    \caption{Results of ResNet-20 on CIFAR-10.
    }
    \label{tb:exp_res20}
	\centering
    {\small
    \begin{tabular}{lccccc}
		\toprule
		{Method} &{No D}  &{No FT}  &{W-bit}  &{A-bit}  &\tabincell{c}{{Top-1}}\\
		\midrule
		Baseline   &--  &-- &32 &32 &94.08  \\
		\midrule
		Real Data  &{\footnotesize\XSolidBrush}   &{\footnotesize\Checkmark} &4  &4  &87.38  \\
		\midrule
		ZeroQ  &{\footnotesize\Checkmark}   &{\footnotesize\Checkmark} &4  &4  &85.39\\
		{DSG} (Ours)  &{\footnotesize\Checkmark}   &{\footnotesize\Checkmark}  &4  &4 &\textbf{87.75}  \\
		\midrule
		Real Data  &{\footnotesize\XSolidBrush}   &{\footnotesize\Checkmark} &6  &6  &93.80  \\
		\midrule
		
		ZeroQ  &{\footnotesize\Checkmark}   &{\footnotesize\Checkmark} &6  &6  &93.33  \\
		{DSG} (Ours)  &{\footnotesize\Checkmark}   &{\footnotesize\Checkmark}  &6  &6 &\textbf{93.79}  \\
		\midrule
		Real Data  &{\footnotesize\XSolidBrush}   &{\footnotesize\Checkmark} &8  &8  &93.95  \\%
		\midrule
		ZeroQ  &{\footnotesize\Checkmark}   &{\footnotesize\Checkmark} &8  &8  &93.94  \\
		{DSG} (Ours)  &{\footnotesize\Checkmark}   &{\footnotesize\Checkmark}  &8  &8 &\textbf{94.07}  \\
		\bottomrule
	\end{tabular}
	
	}
\end{table}

\begin{table}[b]
    \caption{Results of VGG16-bn on CIFAR-10.
    }
   
    \label{tb:exp_vgg16}
        \centering
        {\small
        \begin{tabular}{lccccc}
    		\toprule
    		{Method} &{No D}  &{No FT}  &{W-bit}  &{A-bit}  &\tabincell{c}{{Top-1}}\\
    		\midrule
    		Baseline   &--  &-- &32 &32 &93.86  \\%
    		\midrule
	     	Real Data  &{\footnotesize\XSolidBrush}   &{\footnotesize\Checkmark} &4  &4  &92.50  \\
		    \midrule
		    ZeroQ  &{\footnotesize\Checkmark}   &{\footnotesize\Checkmark} &4  &4  &91.79  \\
	    	{DSG} (Ours)  &{\footnotesize\Checkmark}   &{\footnotesize\Checkmark}  &4  &4 &\textbf{92.89}  \\
    		\midrule
    		Real Data  &{\footnotesize\XSolidBrush}   &{\footnotesize\Checkmark} &6  &6  &93.48  \\%
    		\midrule
    		
    		ZeroQ  &{\footnotesize\Checkmark}   &{\footnotesize\Checkmark} &6  &6  &93.45  \\%
    		{DSG} (Ours)  &{\footnotesize\Checkmark}   &{\footnotesize\Checkmark} &6  &6  & \textbf{93.68} \\%
    		\midrule
    		Real Data  &{\footnotesize\XSolidBrush}   &{\footnotesize\Checkmark} &8  &8  &93.59  \\%
    		\midrule
    		ZeroQ  &{\footnotesize\Checkmark}   &{\footnotesize\Checkmark} &8  &8  &93.53  \\%
    		{DSG} (Ours)  &{\footnotesize\Checkmark}   &{\footnotesize\Checkmark} &8  &8  &  \textbf{93.61}\\%
    		\bottomrule
    	\end{tabular}
    	
    }
\end{table}

\textbf{Results on CIFAR-10 }
We show extra results of our DSG on CIFAR-10~\cite{Filonenko_techreportlearning} dataset with ResNet-20~\cite{7780459} and VGG16-bn~\cite{simonyan2015deep}. See Table~\ref{tb:exp_res20} and~\ref{tb:exp_vgg16}. 
Note that the size of the image sample in the CIFAR-10 dataset is $32\times32$, much smaller than that in the ImageNet dataset ($224\times224$) which is widely evaluated in our paper. The experimental results show that our DSG still outperforms other SOTA generative data-free quantization methods when generating samples with a small size.

\textbf{Evaluation with DFQ } 
DFQ~\cite{Nagel_2019_ICCV} has proposed cross-layer range equalization to equalize the different channel ranges of weight in per-layer quantization and bias correction to eliminate the biased quantization error. Both of the two techniques rely on the statistics of BN layers following the convolution layer.
Therefore, DFQ only works on specific network architectures and cannot be commonly practiced, since BN layers in DFQ are always needed to proceed behind each convolution layer to quantize the corresponding activations. 
Fortunately, generative methods, such as ZeroQ and our DSG, can generate synthetic data for arbitrary architectures, and the statistics of activations can be applied to DFQ replacing the BN statistics. 
Table~\ref{tb:exp_dfq} shows the closeups of two generative data-free quantization method, \ie, ZeroQ and DSG, based on DFQ. Results show that our DSG outperforms ZeroQ by 0.57\% and 3.13\% in W6A6 and W8A8 cases. 

\begin{table}[htb]
    \caption{Evaluation with DFQ using ResNet-18 on ImageNet. We use cross-layer equalization and bias correction proposed by DFQ to perform per-layer quantization.
    }
    
    \label{tb:exp_dfq}
	\centering
	\setlength{\tabcolsep}{2.1mm}
    {\small
    \begin{tabular}{lccccc}
		\toprule
		{Method} &{No D}  &{No FT}  &{W-bit}  &{A-bit}  &\tabincell{c}{{Top-1}}\\
		\midrule
		Baseline   &--  &-- &32 &32 &69.76  \\
		\midrule
		Real Data  &{\footnotesize\XSolidBrush}   &{\footnotesize\Checkmark} &6  &6  &59.16  \\
		\midrule
		ZeroQ  &{\footnotesize\Checkmark}   &{\footnotesize\Checkmark} &6  &6  &58.12 \\
		{DSG} (Ours)  &{\footnotesize\Checkmark}   &{\footnotesize\Checkmark}  &6  &6 &\textbf{58.69}  \\
		\midrule
		Real Data  &{\footnotesize\XSolidBrush}   &{\footnotesize\Checkmark} &8  &8  &69.22  \\%
		\midrule
		ZeroQ  &{\footnotesize\Checkmark}   &{\footnotesize\Checkmark} &8  &8  &65.75  \\
		{DSG} (Ours)  &{\footnotesize\Checkmark}   &{\footnotesize\Checkmark}  &8  &8 &\textbf{68.88}  \\
		\bottomrule
	\end{tabular}
	}
\end{table}

\textbf{More Evaluation with AdaRound }
We present more empirical results on AdaRound~\cite{nagel2020down}, and the experiments can be broadly divided into two categories: quantizing the weight to extremely low bit-width and quantizing both weight and activation. We use image prior~\cite{yin2020dreaming} and labels~\cite{9156818} in these experiments.
First, we quantize the weight to 3/4 bit-width based on the practical lightweight MobileNetV2~\cite{sandler2019mobilenetv2}. As Table~\ref{tb:exp_mbv2} shown, DSG surpasses the SOTA generative methods by 34.33\% with the weight quantized to 3 bit-width. 
Meanwhile, we provide results of DSG with AdaRound on ResNet-18~\cite{7780459} quantized to W4A8 in Table~\ref{tb:exp_activation}, and it also shows that our DSG surpasses ZeroQ by a large margin.

\begin{table}[htb]
    \caption{Evaluation with AdaRound using MobileNetV2 on ImageNet. 
    }
    
    \label{tb:exp_mbv2}
	\centering
	\setlength{\tabcolsep}{1.25mm}
    {\small
    \begin{tabular}{lcccccc}
		
	\toprule
		{Method} &{No D}  &{Label} &{Image Prior}  &{W-bit}  &{A-bit}  &\tabincell{c}{{Top-1}}\\
		\midrule
		Real Data  &{\footnotesize\XSolidBrush} &{\footnotesize\XSolidBrush} 
		&{\footnotesize\XSolidBrush} &3  &32  &58.13 \\
		\midrule
		ZeroQ  &{\footnotesize\Checkmark} &{\footnotesize\Checkmark}  &{\footnotesize\Checkmark} &3  &32  & 11.07 \\
		DSG (Ours)  &{\footnotesize\Checkmark} &{\footnotesize\Checkmark}  &{\footnotesize\Checkmark} &3  &32  & \textbf{45.40} \\
		\midrule
		Real Data  &{\footnotesize\XSolidBrush} &{\footnotesize\XSolidBrush} 
		&{\footnotesize\XSolidBrush} &4  &32  & 68.37\\
		\midrule
		ZeroQ  &{\footnotesize\Checkmark} &{\footnotesize\Checkmark}  &{\footnotesize\Checkmark} &4  &32  &56.16  \\
		DSG (Ours)  &{\footnotesize\Checkmark} &{\footnotesize\Checkmark}  &{\footnotesize\Checkmark} &4  &32  & \textbf{58.13} \\
		\bottomrule
	\end{tabular}
	}
\end{table}

\begin{table}[htb]
    \caption{Evaluation with AdaRound using ResNet-18 on ImageNet. 
    }
    
    \label{tb:exp_activation}
	\centering
	\setlength{\tabcolsep}{1.25mm}
    {\small
    \begin{tabular}{lcccccc}
		
	\toprule
		{Method} &{No D}  &{Label} &{Image Prior}  &{W-bit}  &{A-bit}  &\tabincell{c}{{Top-1}}\\
		\midrule
		Real Data  &{\footnotesize\XSolidBrush} &{\footnotesize\XSolidBrush} 
		&{\footnotesize\XSolidBrush} &4  &8  &68.24 \\
		\midrule
		ZeroQ  &{\footnotesize\Checkmark} &{\footnotesize\Checkmark}  &{\footnotesize\Checkmark} &4  &8  &56.34  \\
		DSG (Ours)  &{\footnotesize\Checkmark} &{\footnotesize\Checkmark}  &{\footnotesize\Checkmark} &4  &8  & \textbf{62.40} \\
		\bottomrule
	\end{tabular}
	}
\end{table}

{
\bibliography{egbib}
\bibliographystyle{ieee_fullname}
}